\documentclass{article} 
\usepackage{iclr2021_conference,times}

\usepackage{amsmath,amsfonts,bm}

\def\eqref#1{equation~\ref{#1}}

\def\1{\bm{1}}

\DeclareMathAlphabet{\mathsfit}{\encodingdefault}{\sfdefault}{m}{sl}
\SetMathAlphabet{\mathsfit}{bold}{\encodingdefault}{\sfdefault}{bx}{n}

\usepackage{hyperref}
\usepackage{url}
\usepackage{booktabs}
\usepackage{subfigure}
\usepackage{graphicx}
\usepackage{wrapfig}

\newcommand{\eref}[1]{Eq. (\ref{#1})}

\title{Hard Encoding of Physics for Learning Spatiotemporal Dynamics}

\author{Chengping Rao, Hao Sun \& Yang Liu\thanks{Corresponding author.} \\
Northeastern University, Boston, MA, USA \\
\texttt{\{rao.che, h.sun, yang1.liu\}@northeastern.edu} \\
}

\iclrfinalcopy 
\begin{document}

\maketitle

\begin{abstract}
Modeling nonlinear spatiotemporal dynamical systems has primarily relied on partial differential equations (PDEs). However, the explicit formulation of PDEs for many underexplored processes, such as climate systems, biochemical reaction and epidemiology, remains uncertain or partially unknown, where very limited measurement data is yet available. To tackle this challenge, we propose a novel deep learning architecture that forcibly encodes known physics knowledge to facilitate learning in a data-driven manner. The coercive encoding mechanism of physics, which is fundamentally different from the penalty-based physics-informed learning, ensures the network to rigorously obey given physics. {\color{black}Instead of using nonlinear activation functions, we propose a novel elementwise product operation to achieve the nonlinearity of the model.} Numerical experiment demonstrates that the resulting physics-encoded learning paradigm possesses remarkable robustness against data noise/scarcity and generalizability compared with some state-of-the-art models for data-driven modeling.
\end{abstract}

\section{Background}

Partial differential equations (PDEs) have played an indispensable role in modeling complex dynamical systems or processes. However, there still exist a considerable portion of dynamical systems, such as those in epidemiology and climate science, whose governing PDEs are unclear or only partially known. To give prediction on systems like these, there have been efforts seeking alternatives to the physics-based models. In recent years, increasing amount of attempts have been made on leveraging physics principles to inform deep neural networks (DNN) for the modeling of systems in a data-driven manner. Among those approaches, the physics-informed neural networks (PINN) is a representative one that can be employed in data-driven modeling \citep{raissi2018deep}, as well as solving forward and inverse PDE problems \citep{raissi2019physics, raissi2019deep, raissi2020hidden, rao2020TAML}. In the framework of PINN, the network is usually informed by physics through a weakly imposed penalty loss consisting of residuals of PDEs and initial/boundary conditions (I/BCs). 

Although PINN has achieved success in modeling dynamical systems, one of its major limitations is that its accuracy relies largely on the these soft physical constraints \citep{wang2020understanding, rao2020physics} which may not be satisfied well during training due to the ill-posedness of the optimization problem. Furthermore, the use of fully connected layers poses intrinsic limitations to low-dimensional parameterizations. Efforts have been placed to overcome these issues by employing discrete learning schemes via convolutional filters, such as HybridNet \citep{long2018hybridnet}, dense convolutional encoder-decoder network \citep{zhu2019physics}, auto-regressive encoder-decoder model \citep{geneva2020modeling}, TF-Net \citep{wang2020towards}, DiscretizationNet \citep{ranade2021discretizationnet} and PhyGeoNet \citep{gao2021phygeonet}. These methods generally show better computational efficiency and accuracy. However, the core learning component of these networks is still a black box and the resulting models lack the capability to ``hard-encode'' our prior physical knowledge. How to construct a data-driven model that both fits limited data and generalizes well the underlying physics remains a critical challenge. 

To address this challenge, in this work, we propose a physics-encoded recurrent-convolutional neural network (PeRCNN), which forcibly encodes the physics structure to facilitate learning for data-driven modeling of nonlinear systems. The physics-encoding mechanism guarantees the model to rigorously obey the given physics based on our prior knowledge. Instead of using activation functions, which results in poor interpretability and generalizability, we achieve nonlinear approximation via elementwise product among the feature maps, leading to a recurrent $\Pi$-block that renders PeRCNN with good expressiveness and flexibility at representing complex nonlinear physics. The $\Pi$-block mimics governing terms in a PDE. The spatial dependency is learned by either convolutional or predefined finite-difference-based filters while the temporal evolution is modeled by a forward Euler time marching scheme. Numerical experiments demonstrate that PeRCNN outperforms the existing models (e.g., ConvLSTM, ResNet and DHPM) in the metrics of generalizability.

Our work is closely related to the deep residual network (ResNet) which addresses the notorious problem of gradient vanishing/exploding for very deep networks with residual learning \citep{he2016deep}. Our network architecture also features the residual connection that enables the residual learning of dynamical systems, which has been interpreted by others as the forward Euler time-stepping scheme \citep{chen2015learning, chang2017multi, chen2018neural, ruthotto2019deep}. People have employed the recurrent ResNet, a variant of ResNet whose parameters are shared across time, to solve spatiotemporal prediction problems \citep{liao2016bridging, zhang2017deep}. Although ResNet has shown success in lots of applications, residual blocks composed of traditional convolutional or fully connected layers still face the issue of poor interpretability, hence hindering its applications to spatiotemporal dynamical systems where governing PDEs are potentially available.

\section{Methodology}
\label{gen_inst}

In this part, we elaborate the principle for designing the network architecture of PeRCNN for data-driven modeling of spatiotemporal dynamical systems. The formulation to establish a data-driven model from limited and noisy measurements is introduced in \ref{sec:formula_data_driven} with more details. 

\subsection{Network architecture}\label{sec:architecture}

\begin{wrapfigure}[25]{r}{0.53\textwidth}
\vspace{-5pt}
\begin{center}
\includegraphics[width=0.99\linewidth]{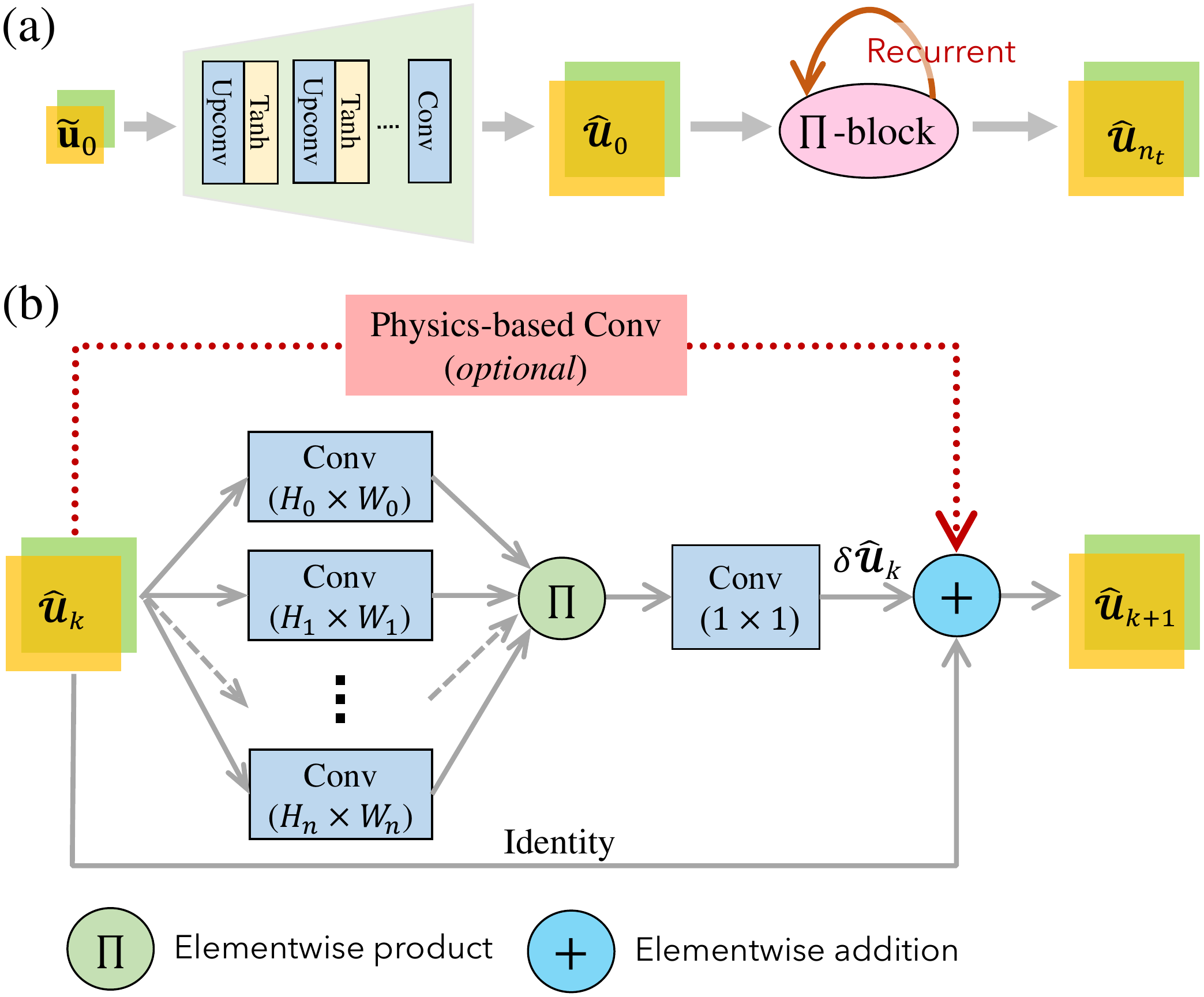}
\vspace{-20pt}
\caption{Schematic architecture of PeRCNN: (a) the network with a recurrent $\Pi$-block folded; and (b) $\Pi$-block for the recurrent computation. Here, $\tilde{\mathbf{u}}_0$ is the low-resolution noisy initial state measurement, while $\hat{\boldsymbol{\mathcal{U}}}_k$ is the predicted full-resolution solution at time $t_k$. The decoder (initial state generator) is used to downscale/upsample the low-resolution initial state.}
\label{Diagram}
\end{center}
\vspace{-12pt}
\end{wrapfigure}

Our proposed PeRCNN (see Fig. \ref{Diagram}) consists of two major components: a fully convolutional (Conv) network as initial state generator (ISG) and an unconventional Conv block, namely $\Pi$-block (product), for recurrent computation, as depicted in Fig. \ref{Diagram}(a). ISG enables a mapping from the noisy low-resolution measurement $\Tilde{\mathbf{u}}_0$ to the full-resolution initial state $\hat{\boldsymbol{\mathcal{U}}}_0$ from which the recurrent computation can be initiated. 

In the $\Pi$-block shown in Fig. \ref{Diagram}(b), which is the core of PeRCNN, the state variable $\hat{\boldsymbol{\mathcal{U}}}_k$ from the previous time step first goes through multiple parallel Conv layers, whose feature maps will then be fused via an elementwise product layer. A $1\times 1$ Conv layer (or network in network \citep{lin2013network}) is appended after the product operation to aggregate multiple channels into the output of desired number of channels. Assuming the output of the $1\times1$ Conv layer approximates the nonlinear function $\mathcal{F}(\cdot)$, we multiply it by the time spacing $\delta t$ to obtain the residual of the dynamical system at time $t_k$, i.e., $\delta \hat{\boldsymbol{\mathcal{U}}}_k$. The $\Pi$-block operation is expressed as: 
\begin{equation} 
    \label{eq:update_rule} 
    \hat{\boldsymbol{\mathcal{U}}}_{k+1}=\hat{\boldsymbol{\mathcal{U}}}_k +\left \{ \left [ \prod_{i=0}^{n} \left ( \hat{\boldsymbol{\mathcal{U}}}_k\ast \mathbf{W}_i+\mathbf{b}_i \right )\right ] \ast \mathbf{W}^{(1)} + \mathbf{b}^{(1)} \right \} \delta t
\end{equation}
where $\ast$ denotes the Conv operation; $\mathbf{W}_i$ and $\mathbf{b}_i$ are the weight and bias for the filter in the $i^\text{th}$ layer while $\mathbf{W}^{(1)}$ and $\mathbf{b}^{(1)}$ correspond to those of the $1\times1$ Conv filter; $n+1$ is the number of parallel Conv layers; $\Pi$ denotes the elementwise product; $\delta t$ is the selected time spacing. It should be noted that a highway physics-based Conv layer (see Fig \ref{Diagram}(b)) could be created when some specific terms are known \textit{a priori} in the PDEs. Such a highway connection is not necessary to the succeed of data-driven modeling but could accelerate the training speed and improve the extrapolation accuracy.

It is worth noting that we achieve the nonlinearity of our network via elementwise product of the feature maps instead of using nonlinear activation functions, mainly for three reasons: \textbf{(1)} Though the nonlinear activation function is crucial to the expressiveness of the DL model, it is also a source of poor interpretability. We consider it unfavorable to use these nonlinear functions to build a recurrent block that aims to generalize the unknown physics. \textbf{(2)} The nonlinear function $\mathcal{F}$ in the form of polynomial\footnote{Polynomial herein encompasses linear derivative terms, e.g., $\mathbf{u}\cdot\nabla u+u^2v$ has 2$^\text{nd}$ and 3$^\text{rd}$ order terms.} covers a wide range of well-known dynamical systems, such as Navier-Stokes, reaction-diffusion (RD), Lorenz, Schr$\ddot{\text{o}}$dinger equations, to name only a few. Since the spatial derivatives can be computed by Conv filters \citep{cai2012image}, a $\Pi$-block with $n$ parallel Conv layers of appropriate filter size is able to represent a polynomial up to the $n^\text{th}$ order. \textbf{(3)} Compared with the regression models that heavily rely on predefined basis functions \citep{brunton2016discovering}, the $\Pi$-block is flexible at generalizing the function $\mathcal{F}$. For example, a $\Pi$-block with 2 parallel layers of suitable filter size ensembles a family of polynomials up to 2$^\text{nd}$ order (e.g., $u$, $\Delta u$, $uv$, $\mathbf{u}\cdot\nabla u$), with no need to explicitly define the basis. 

Although we mainly consider the nonlinear function $\mathcal{F}$ in the form of polynomial, terms of other forms such as trigonometric and exponential functions can be incorporated by adding a particular symbolic activation (e.g., sin, cos, exp, etc.) layer following the Conv operation. 

\subsection{Hard encoding mechanism of physics}

The encoding mechanism is employed to strictly impose the prior physical knowledge of the system to PeCRNN, which contributes to a well-posed optimization problem. In this work, two types of physics can be considered for hard-encoding, namely, the prior knowledge on I/BCs and active terms in the governing PDEs. The ICs (or initial states) can be naturally imposed when PeRCNN starts the recurrent computation from $\tilde{\mathbf{u}}_0$. For the BCs (Dirichlet or Neumann type), we borrow the idea from the finite difference (FD) method and apply the physics-based padding to the model's prediction at each time step (i.e., $\hat{\boldsymbol{\mathcal{U}}}_k$). More specifically, we pad the prediction with prescribed values defined by the Dirichlet BCs. The padding operations of the Neumann BCs will be computed based on the boundary values and their gradient information. 


PeRCNN also has the capability to encode prior-known terms in the governing PDEs via a highway Conv layer (see Fig. \ref{Diagram}(b)) with predefined FD-based filters. In Section \ref{sec:experiment}, we consider a reaction-diffusion system and assume the diffusion term $\Delta\mathbf{u}$ is known. Therefore, a Conv layer with discrete Laplacian operator (see \eref{eq:diff_spatial_filter}) as its filter is created to approximate $\Delta\mathbf{u}$. The associated diffusion coefficient is unknown and placed as part of the trainable variables. It should be noted that, by using the residual connection in the recurrent $\Pi$-block, we also implicitly encode the existing term of $\mathbf{u}_t$.

\section{Experiments}
\label{sec:experiment}

\subsection{Datasets}

{\color{black}In the experiments, we employ two different dynamical systems, 2D Burgers' equation and 3D Gray-Scott (GS) reaction-diffusion (RD) equation, to examine our approach. The 2D Burgers' equation, with a wide applications in applied mathematics, such as fluid/traffic flow modeling, is given by 
\begin{equation}
    \label{eq:burgers}
    \mathbf{u}_t+ \mathbf{u} \cdot \nabla \mathbf{u} = \nu \Delta \mathbf{u}
\end{equation}
where $\mathbf{u}=[u,~v]^\texttt{T}$ denotes the fluid velocities; $\nabla$ is the Nabla operator; $\Delta$ is the Laplacian operator and $\nu$ is the viscosity coefficient selected to be 0.005 in this case. We consider a computational domain of $\Omega\times[0,T]=[-0.5,0.5]^2\times[0,0.4]$ under periodic boundary conditions and generate the solution on a $101^2\times 1601$ spatiotemporal grid. }

The second dataset considered is the 3D GS-RD equation, which can be described by
\begin{equation} 
    \label{eq:gs_eqn} 
    \mathbf{u}_t=\mathbf{D}\Delta \mathbf{u}+\mathbf{R(u)} 
\end{equation}
Here, $\mathbf{u}=[u,~v]^\texttt{T}$ is the concentration vector; $\mathbf{D}=\text{diag}(\mu_u,\mu_v)$ is the diagonal diffusion coefficient matrix; $\mathbf{R(u)}=[-uv^2+f(1-u),~uv^2-(f+\kappa)v]^\texttt{T}$ is the nonlinear reaction vector, where $\kappa$ and $f$ denote the kill and feed rate, respectively. This model has found wide applications in computational chemistry and biochemistry. In our dataset, the parameters of $\mu_u=0.2,~\mu_v=0.1,~\kappa=0.055$ and $f=0.025$ are employed. We consider the physical domain of $\Omega\times[0,T]=[-50,50]^3\times[0,750]$ with periodic boundary condition and generate the solution on a Cartesian grid ($49^3\times 1501$) using the FD method. For both of the two cases, we assume there is diffusion phenomenon observed in the system, i.e., $\Delta \mathbf{u}$ will appear in the governing PDE. In addition, 9-point stencil (see \eref{eq:diff_spatial_filter} in \ref{sec:laplacian}) is used to compute the diffusion terms while the Runge-Kutta method is adopted for time stepping. The loss function employed in the training and the evaluation metrics are detailed in \ref{sec:loss} and \ref{sec:aRMSE}.

\subsection{Results}
To evaluate the performance of the proposed PeRCNN, we make a comparison of the prediction between the PeRCNN and some widely used spatiotemporal predictive models, e.g., ConvLSTM \citep{shi2015convolutional}, recurrent ResNet \citep{ liao2016bridging, zhang2017deep} and the deep hidden physics model (DHPM) \citep{raissi2018deep}. A brief introduction to each method and the range of hyperparameters considered are given in \ref{sec:baseline}. 

\subsubsection{2D Burgers' dataset}\label{sec:2d_burgers}
\begin{figure}
    \centering
    \begin{minipage}{0.48\textwidth}
        \vspace{2pt}
        \centering
    	\includegraphics[width=1.0\linewidth]{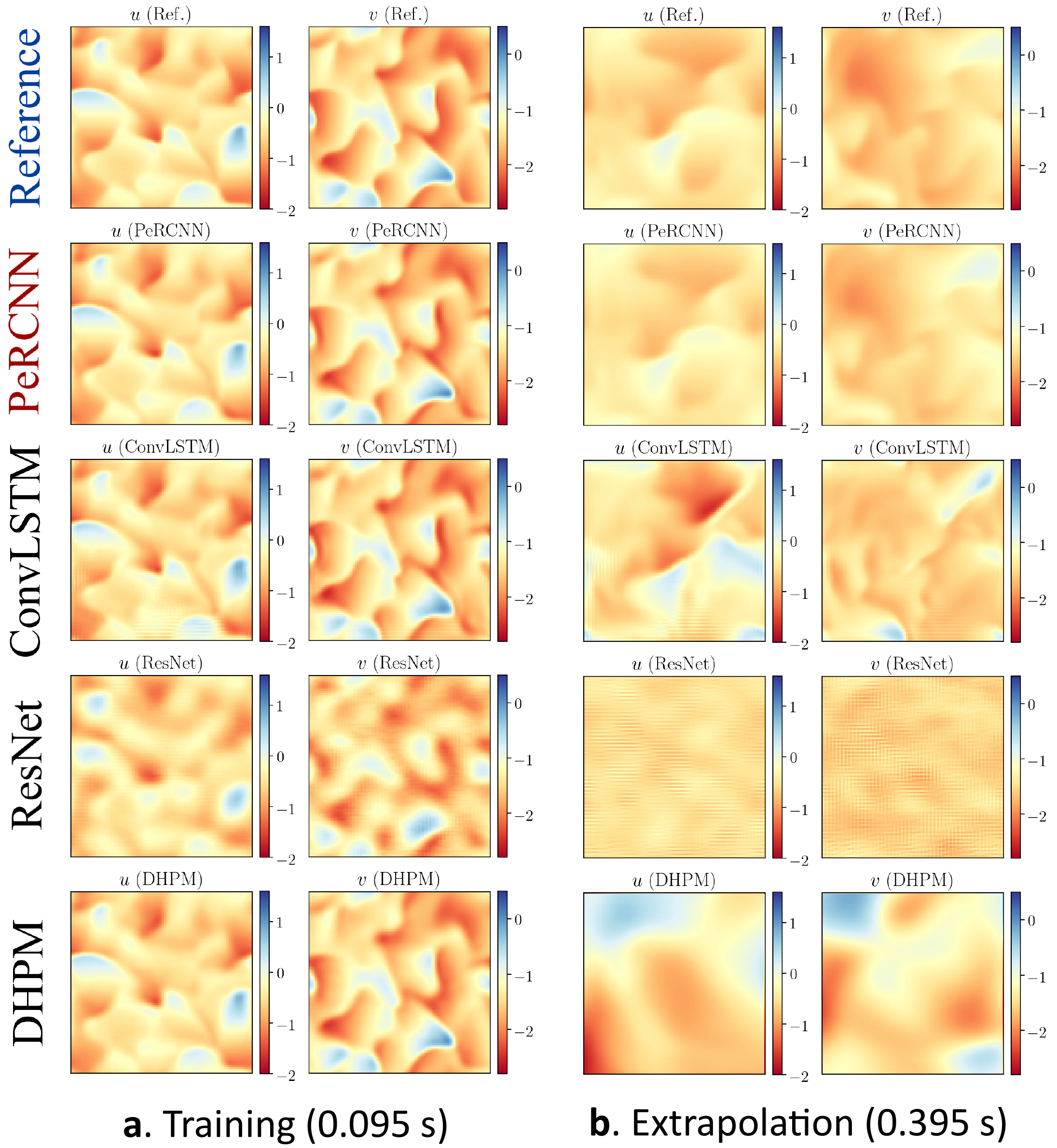} 
    	\caption{Contours for 2D Burgers' predictions.}
    	\label{2D_BG_snapshots}
    \end{minipage}\hfill
    \begin{minipage}{0.48\textwidth}
        \centering
    	\includegraphics[width=1.0\linewidth]{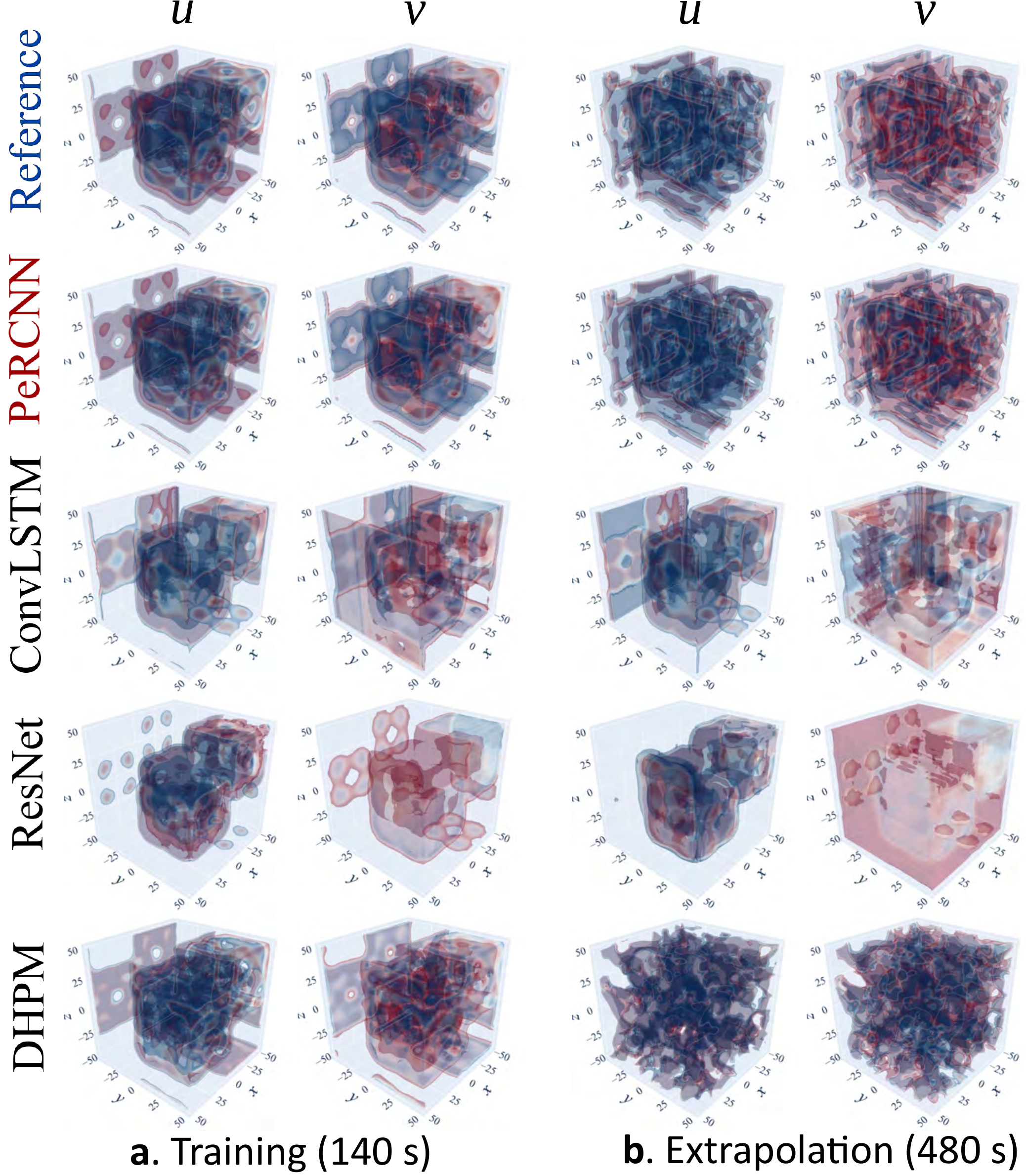}
    	\caption{Isosurfaces for 3D GS-RD predictions. ($u$: blue = 0.5, red = 0.3; $v$: blue = 0.3, red = 0.1). \vspace{-12pt}}
    	\label{3D_GS_snapshots}
    \end{minipage}
\end{figure}

{\color{black}In this case, the training data includes 11 low-resolution ($51\times 51$) snapshots uniformly selected from the time interval of $[0,0.1]$ after 10\% Gaussian noise is added to the original dataset. Also, 2 snapshots are used as the hold-out validation dataset for hyperparameters selection and early stopping. Each model is constructed to produce the prediction with full spatiotemporal resolution, i.e., $\hat{\boldsymbol{\mathcal{U}}}\in\mathbb{R}^{2\times 401\times 101\times 101}$. After the model is finalized\footnote{All the implementations are coded in PyTorch or TensorFlow on a NVIDIA Tesla v100 GPU (32G).}, 1200 extra prediction steps are performed to evaluate how the learned model generalizes beyond the training regime. 
Fig. \ref{2D_BG_snapshots} shows the snapshots predicted by each model at $t=0.095$ and $0.395$. As shown in Fig. \ref{2D_BG_snapshots}(a), all the models are able to fit the training data and produce satisfactory prediction in the supervised time period. However, when it comes to long-term extrapolation, the model predictions deviate from the ground truth significantly except for PeRCNN, which demonstrates that PeRCNN generalizes the unknown underlying physics well from the data. This conclusion is further confirmed by Fig. \ref{error_prop}(a) which depicts the accumulative RMSE (see \eref{accum_rmse}). We may notice that the accumulative RMSE starts from an initial high value. This is due to the fact that the training data is corrupted by 10\% Gaussian noise and the metrics is computed from one single snapshot at the beginning. The effect of the unrelated noise gradually fades out as more time steps are considered.}

\subsubsection{3D Gray-scott RD dataset}

\begin{figure*}[t!]
\centering
\subfigure[2D Burgers' dataset]{
\begin{minipage}[t]{0.45\linewidth}
\centering
\includegraphics[width=\linewidth]{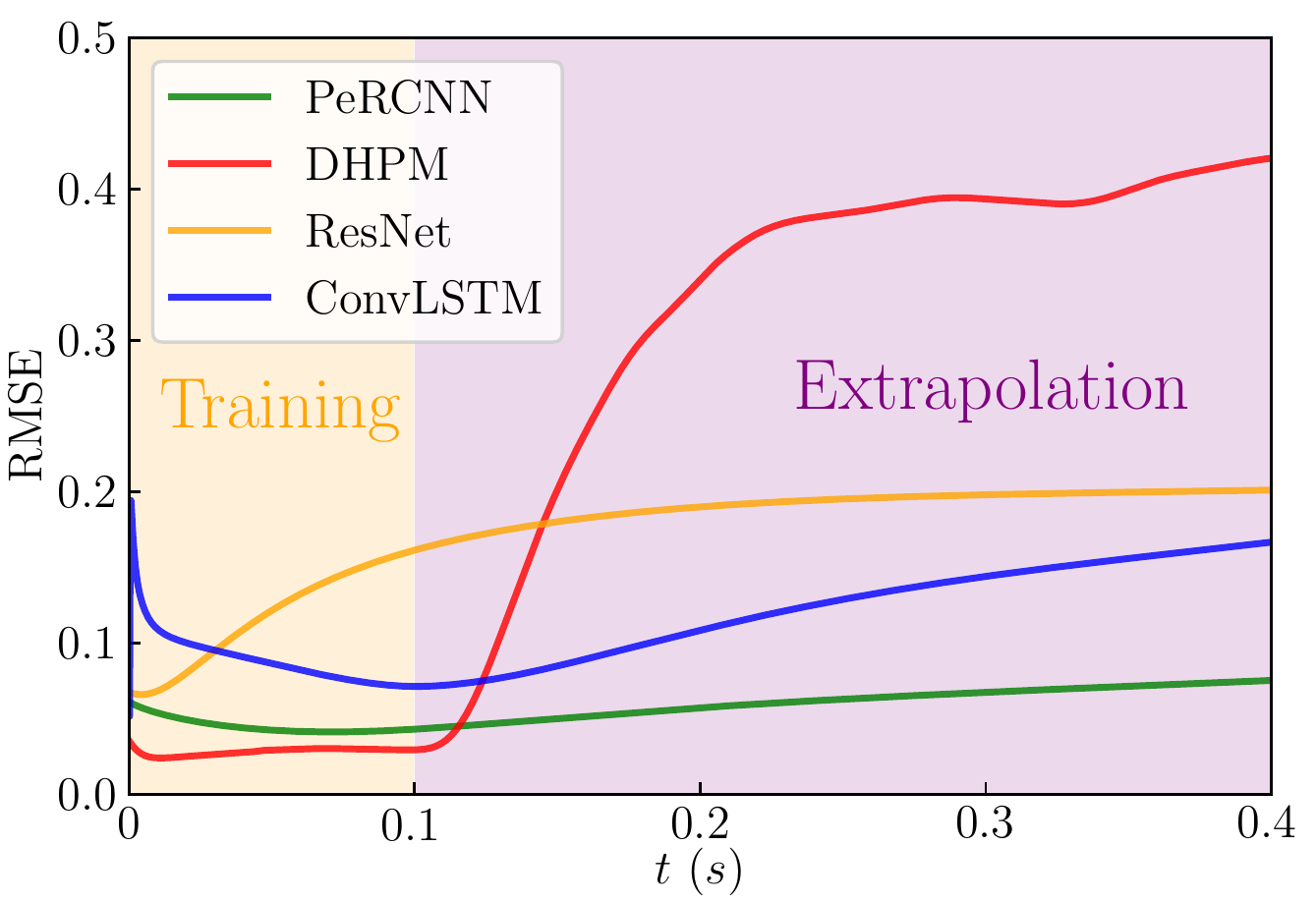}
\end{minipage}%
}%
\hfill
\centering
\subfigure[3D GS-RD dataset]{
\begin{minipage}[t]{0.45\linewidth}
\centering
\includegraphics[width=\linewidth]{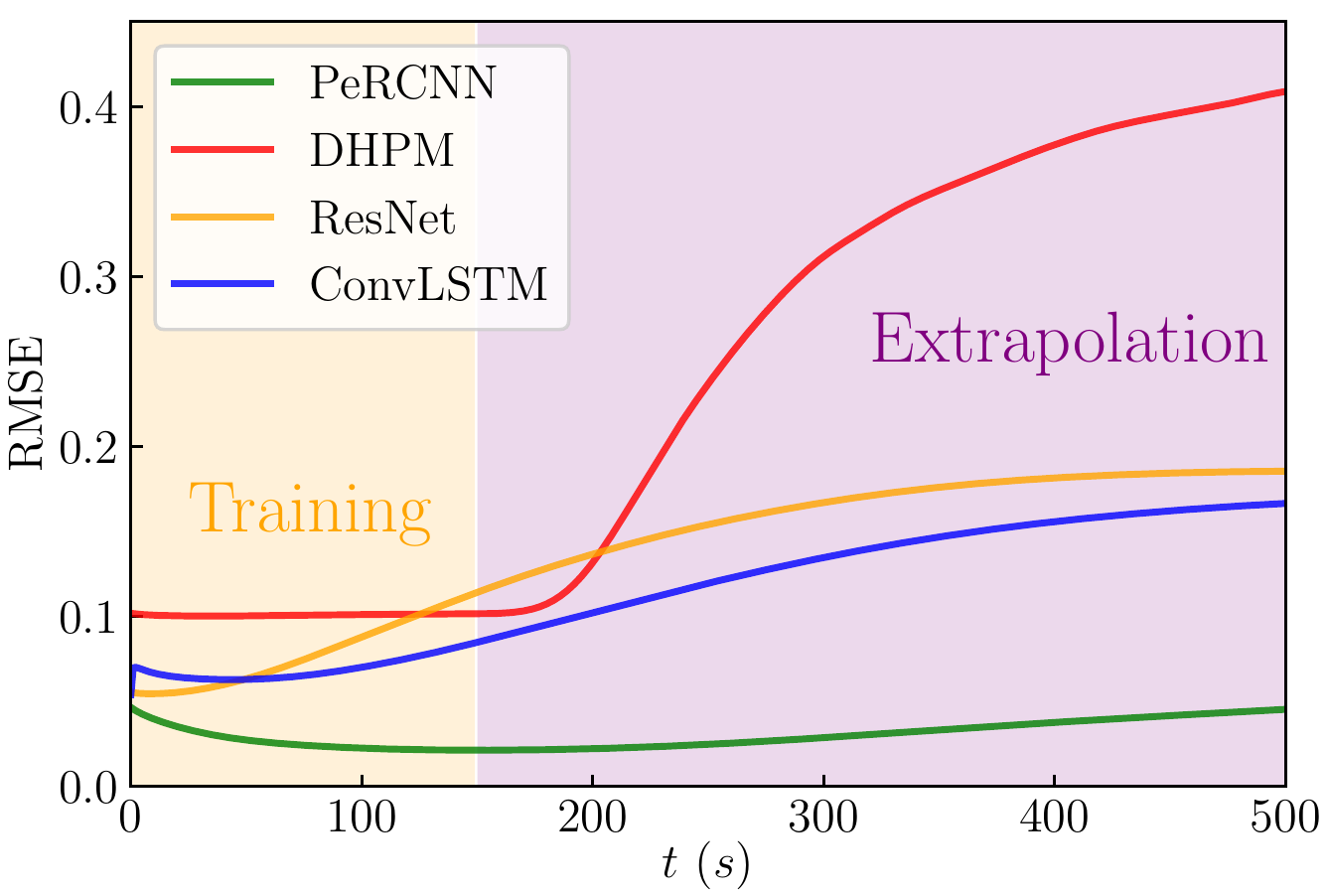}
\end{minipage}
}%
\vspace{-10pt}
\caption{Error propagation of the training and extrapolation prediction.}
\label{error_prop}
\end{figure*}


In this example, the training data includes 21 noisy snapshots on a $25^3$ grid sampled from $t=0$ to $150$. Each trained model produces 301 full-resolution snapshots during supervised learning stage while 700 extrapolation steps are predicted after each model is finalized. 

The predicted isosurfaces of two levels are plotted in Fig. \ref{3D_GS_snapshots}. It can be observed that most of the models are able to produce reasonable prediction during the supervised time period. However, for the time beyond the training regime, the PeRCNN outperforms remaining models significantly. The flat error propagation curve of PeRCNN, as shown in Fig. \ref{error_prop}(b), also demonstrates the remarkable generalization capability of PeRCNN. 

{\color{black}As explained in Section \ref{sec:architecture}, the architecture of PeRCNN enables a better approximation to the $\mathcal{F}$ of nonlinear PDEs, resulting in the good generalization capability of our model. To verify this claim, we exploit the multiplicative form of the $\Pi$-block and extract the explict expression from the learned model. We find the equivalent expression of the learned model is quite close to the genuine PDE. For more details on interpreting the learned PeRCNN model, please refer to Section \ref{sec:interpret}. }


\section{Conclusions}
A novel DL architecture, called PeRCNN, is developed for modeling of nonlinear spatiotemporal dynamical systems based on sparse and noisy data. Our prior physics knowledge is forcibly encoded into PeRCNN which guarantees the resulting network strictly obeys given physics (e.g., I/BCs or known terms in PDEs). This brings distinct benefits for improving the convergence of training and accuracy of the model. To evaluate the generalizability of PeRCNN, the trained model is used for extrapolation along the temporal horizon. Comparisons with several state-of-the-art models demonstrate that physics-encoded learning paradigm uniquely possesses remarkable robustness against data noise/scarcity and generalizability. {\color{black}Equally important, PeRCNN shows good interpretability due to the multiplicative form of the recurrent block. As shown in Section \ref{sec:interpret}, an explicit expression can be extracted from the learned model via some symbolic computations.} 

Although PeRCNN shows promise in data-driven modeling of complex systems, it is restricted by the computational bottleneck due to the high dimensionality of the discretized system, especially when it comes to systems in a large 3D spatial domain with long-term evolution. However, this issue is expected to be addressed via temporal batching and multi-GPU training, to be investigated in our future studies.




\newpage

\bibliography{iclr2021_conference}
\bibliographystyle{iclr2021_conference}

\appendix
\section{Appendices}\label{sec:appd}

\subsection{Discrete Laplacian operator}\label{sec:laplacian}

The discrete Laplacian operator, defined by \eref{eq:diff_spatial_filter}, is used in finite difference (FD) method to generate the synthetic dataset and in the PeRCNN to create physics-based convolutional layers. 

\begin{equation}
    \label{eq:diff_spatial_filter}
    W_\Delta = \frac{1}{12(\delta x)^2} {
    \begin{bmatrix} 
        0 & 0 & -1 & 0 & 0 \\
        0 & 0 & 16 & 0 & 0 \\
        -1 & 16 & -60 & 16 & -1 \\
        0 & 0 & 16 & 0 & 0 \\  
        0 & 0 & -1 & 0 & 0 
    \end{bmatrix}}
\end{equation}

\subsection{Formulation of data-driven modeling}\label{sec:formula_data_driven}
Let us consider a spatiotemporal dynamical system described by a set of nonlinear, coupled PDEs, expressed as
\begin{equation} 
    \label{dynamic_system_pde} 
    \mathbf{u}_t=\mathcal{F}\left(\mathbf{x}, t, \mathbf{u}, \mathbf{u}^2, \nabla_\mathbf{x}  \mathbf{u}, \mathbf{u}\cdot\nabla_\mathbf{x} \mathbf{u}, \nabla^2 \mathbf{u}, \cdots\right)
\end{equation}
where the state variable/solution $\mathbf{u}(\mathbf{x},t)\in\mathbb{R}^s$ (e.g., $\mathbf{u}=[u,~v]^\texttt{T}$ for $s=2$) is defined over the spatiotemporal temporal domain $\{\mathbf{x}, t\} \in \Omega\times[0,T]$; $\nabla_{\mathbf{x}}$ is the Nabla operator with respect to $\mathbf{x}$; and $\mathcal{F}(\cdot)$ is a nonlinear function. The solution to this problem is subject to the initial condition (IC)  $\mathcal{I}(\mathbf{u},\mathbf{u}_t;t=0,\mathbf{x} \in \Omega)=0$ and boundary condition (BC) $\mathcal{B}(\mathbf{u},\nabla_{\mathbf{x}}\mathbf{u}, \cdots;\mathbf{x} \in \partial \Omega)=0$, where $\partial \Omega$ denotes the boundary of the system. In this work, we mainly focus on regular physical domains, i.e., $\mathbf{u}$ can be discretized on a $H\times W$ Cartesian
grid at time steps $\{t_1,..., t_k, ..., t_{n_t}\}$, where $n_t$ denotes the total number of time steps. Provided a scarce and potentially noisy set of measurements $\Tilde{\mathbf{u}}\in\mathbb{R}^{2\times n_t'\times H'\times W'}$ over a coarser spatiotemporal grid, the goal of the data-driven modeling is to establish a reliable model that gives the most likely full-field solution $\boldsymbol{\hat{\mathcal{U}}}\in \mathbb{R}^{2\times n_t\times H \times W}$ and possesses satisfactory extrapolation ability over the temporal horizon (e.g., for $t > t_{n_t}$). 

\subsection{Baselines for comparisons}\label{sec:baseline}

To make a comparison of the PeRCNN with other models used widely for data-driven modeling of spatiotemporal systems, we also implemented the recurrent ResNet \citep{liao2016bridging, zhang2017deep}, convolutional long-short term memory (ConvLSTM) \citep{shi2015convolutional} and deep hidden physics model (DHPM) \citep{raissi2018deep}. A very brief introduction to each method is given below for readers to grasp the major characteristics of each model.

\textbf{ConvLSTM} \citep{shi2015convolutional} is a convolutional variant of LSTM which exploits multiple self-parameterized controlling gates, such as input, forget and output gates, to capture the spatiotemporal correlations among the data. It has been extensively used in applications such as video super-resolution \citep{tao2017detail,liang2017dual}, traffic prediction \citep{yuan2018hetero} and climate forecasting \citep{shi2015convolutional}, among many others. 

\textbf{Recurrent ResNet} is another model adopted widely by researchers \citep{ liao2016bridging, zhang2017deep} for the spatiotemporal prediction of dynamical systems. One main characteristic distinguishes the recurrent ResNet with the conventional ResNet \citep{he2016deep} is that the weights are shared across time.

\textbf{DHPM} \citep{raissi2018deep} differs from the previous two models as it utilizes fully connected neural networks (FCNNs) and exert hidden (unknown) physics prior on the solution. In DHPM, one deep FCNN is employed to fit the data, i.e., pairs of the spatiotemporal location and solution, while another shallow FCNN is used to impose a hidden physical constraint on the fitted solution. 

Since it is impossible to keep the hyperparameters exactly the same among different models, we select the best configuration for each model from a range of hyperparameters, which are summarized in Table \ref{tb:hyerpara_range_percnn}, \ref{tb:hyerpara_range_convlstm}, \ref{tb:hyerpara_range_resnet} and \ref{tb:hyerpara_range_dhpm}. In addition to the listed hyperparameters, all other hyperparameters are kept the same, e.g., training/validation/testing dataset split, the number of prediction steps, the optimizer (Adam), the max number of epochs, the Gaussian noise level (10\%) and the random seed. In the network architecture design, we assume the solution within the domain is periodic while the dynamical system of interest is accompanied with the ubiquitous diffusion phenomenon. Hence, a diffusion Conv layer with fixed filters will be created in the following PeRCNN models.

\begin{table*}[h!]
\begin{center}
\caption{Range of hyperparameters for PeRCNN.}
\label{tb:hyerpara_range_percnn}
\begin{tabular}{lcccccc}
\toprule 
Dataset & Filter size & $\#$ layers & $\#$ channels  & $\#$ channels & Learning Rate & $\lambda$  \\
  &  &  ($\Pi$-block)  &(ISG) &  &  \\
\midrule
2D BE & 1$\sim$5 (5) & 2$\sim$4 (4) & 4$\sim$16 (8) & 4$\sim$16 (8) & 0.001$\sim$0.01 (0.002)& 0.001$\sim$1 (1) \\
3D GS & 1$\sim$5 (1) & 2$\sim$4 (3) & 2$\sim$8 (4) & 4$\sim$8 (4) & 0.001$\sim$0.01 (0.005) & 0.001$\sim$1 (0.5) \\
\bottomrule
\end{tabular}
\end{center}
\vspace{-10pt}
\end{table*}

\begin{table*}[h!]
\begin{center}
\caption{Range of hyperparameters for ConvLSTM.}
\label{tb:hyerpara_range_convlstm}
\begin{tabular}{lcccccc}
\toprule 
Dataset & Filter size & $\#$ layers & $\#$ channels & Learning rate & Weight decay   \\
\midrule
2D BE & 3$\sim$5 (5) & 1$\sim$2 (2) & 16$\sim$32 (32) & 0.0005$\sim$0.01 (0.001)& e-5$\sim$e-3 (e-5) \\
3D GS & 3$\sim$5 (5)  & 1   & 8$\sim$16 (16) & 0.0005$\sim$0.01 (0.0005) & e-5$\sim$e-3 (e-5)  \\
\bottomrule
\end{tabular}
\end{center}
\vspace{-10pt}
\end{table*}

\begin{table*}[h!]
\begin{center}
\caption{Range of hyperparameters for recurrent ResNet.}
\label{tb:hyerpara_range_resnet}
\begin{tabular}{lcccccc}
\toprule 
Dataset & Filter size & $\#$ layers & $\#$ channels & Learning rate & Weight decay   \\
\midrule
2D BE & 3$\sim$5 (3) & 2$\sim$4 (2) & 16$\sim$128 (64) & 0.0005$\sim$0.01 (0.002)& e-5$\sim$e-2 (e-4) \\
3D GS & 3$\sim$5 (3)  & 2$\sim$3 (2)  & 8$\sim$32 (32) & 0.0005$\sim$0.01 (0.001) & e-5$\sim$e-2 (e-4)  \\
\bottomrule
\end{tabular}
\end{center}
\vspace{-10pt}
\end{table*}

\begin{table*}[h!]

\begin{center}
\caption{Range of hyperparameters for DHPM.}
\vspace{-10pt}
\label{tb:hyerpara_range_dhpm}
    \resizebox{1.0\linewidth}{!}{
        \begin{tabular}{lcccccc}
        \toprule 
        Dataset & $\mathcal{N}_1$ width & $\mathcal{N}_1$ depth & $\mathcal{N}_2$ width & $\mathcal{N}_2$ depth & Input for $\mathcal{N}_2$ & Learning rate    \\
        \midrule
        2D BE & 80$\sim$120 (120) & 4$\sim$5 (5) & 10$\sim$30 (20) & 2$\sim$3 (2) & $(\Delta u,\Delta v,uu_x,$ & 0.001 $\sim$0.02 (0.005) \\
        &&&&&$vu_y,uv_x,vv_y)$&\\
        3D GS & 60$\sim$100 (80)  & 4$\sim$5 (5) & 10$\sim$30 (10) & 2$\sim$3 (2)& $(\Delta u,\Delta v,u,v)$ & 0.001 $\sim$0.02 (0.01)  \\
        \bottomrule
        \end{tabular}
    }
\end{center}
\vspace{-10pt}
\end{table*}

\subsection{Loss function}\label{sec:loss}

Given the low resolution measurement\footnote{It can be readily generalized to 3D cases.} $\Tilde{\mathbf{u}}\in\mathbb{R}^{2\times n_t'\times H'\times W'}$ where $n_t'<n_t$, $H'<H$ and $W'<W$, our goal of the data-driven modeling is to reconstruct the most likely full-field solution $\boldsymbol{\hat{\mathcal{U}}}\in \mathbb{R}^{2\times n_t\times H \times W}$. The loss function to train PeRCNN is defined as:
\begin{equation} 
    \label{sys_id_loss} 
    \displaystyle
    \mathcal{L}(\mathbf{W}, \mathbf{b})=\textrm{MSE}\left(\boldsymbol{\hat{\mathcal{U}}}(\tilde{\mathbf{x}})-\tilde{\mathbf{u}}\right)+\lambda\cdot \textrm{MSE}\left(\boldsymbol{\hat{\mathcal{U}}}_0-\mathcal{P}(\tilde{\mathbf{u}}_0)\right)
\end{equation}
where $\boldsymbol{\hat{\mathcal{U}}}(\tilde{\mathbf{x}})$ denotes the network's prediction at the coarse grid nodes $\tilde{\mathbf{x}}$; $\tilde{\mathbf{u}}$ denotes the low-resolution measurement;  $\mathcal{P}(\cdot)$ is a spatial interpolation function (e.g., bicubic or bilinear); $\lambda$
is the weighting coefficient for the regularizer. The regularization term denotes the IC discrepancy between the interpolated initial state and the network's prediction, which is found effective in preventing network overfitting.

\subsection{Evaluation metrics}\label{sec:aRMSE}

Accumulative rooted-mean-square error (RMSE), defined by \eref{accum_rmse}, is employed to compute the error of all snapshots before a time step $t_k$. It is used throughout this paper to evaluate the error propagation of the model prediction. 

\begin{equation} 
    \label{accum_rmse} 
    \begin{aligned}
    \text{RMSE}(t_k) = \sqrt{\frac{1}{nk}\sum_{i=1}^{k}\left \|\hat{\boldsymbol{\mathcal{U}}}_i-\boldsymbol{\mathcal{U}}^{\text{ref}}_i\right \|_2^2}
    \end{aligned}
\end{equation}

where $\boldsymbol{\mathcal{U}}^{\text{ref}}_i$ is the reference solution and $k\in\{1,2,\cdots~n_t\}$.

{\color{black}
\subsection{Interpret the learned model}\label{sec:interpret}

Since each channel of the input ($\hat{\boldsymbol{\mathcal{U}}}_k$) denotes a state variable component (i.e., $[u,v]$), the multiplicative form of $\Pi$-block makes it possible to extract (or interpret) an explicit expression for $\mathcal{F}$, in a symbolic way, from the learned weights and biases. We first interpret the learned model from 3D GS-RD case, whose parallel Conv layers have filter size of 1. In this case, each output channel from the Conv layer would be the linear combination of $u$, $v$ and a constant. The identified diffusion coefficient matrix from the physics-based Conv layer is $\mathbf{D}=\text{diag}(0.18,0.080)$. The extracted expression from the learned $\Pi$-block is
\begin{equation}
    \label{eq:factor_out_terms}
    \resizebox{0.99\linewidth}{!}{$
    \mathbf{R(u)}=
    \begin{bmatrix}
    -0.0074u^3 - 0.0051u^2v - 0.2uv^2 - 0.0386v^3 - 0.0018u^2 - 0.11uv- 0.055v^2 - 0.016u  - 0.022v + 0.025 \\
    0.0005u^3 - 0.013u^2v + 0.54uv^2 - 0.087v^3 - 0.0076u^2 + 0.023uv + 0.046v^2 + 0.017u - 0.036v - 0.0097
    \end{bmatrix} 
    $}
\end{equation}
which includes some distracting terms with small coefficients, due to the 10\% noise and scarcity of the training data. 

To interpret terms involving partial derivatives (e.g., $u\nabla^2u$, $uu_x$), it would require us to completely freeze or impose moment matrix constraints on part of the convolutional filters \citep{long2018pde}. Here, a simple experiment on 2D Burgers' equation is conducted. The network employed has two Conv layers with two channels. The first Conv layer are associated with $\partial x$ and $\partial y$ respectively, by fixing the filters with corresponding FD stencils. The remaining settings are kept the same as in Section \ref{sec:2d_burgers} except that noise-free training data is used. The interpreted expression from the whole PeRCNN model is
\begin{equation}
    \label{eq:burgers_terms}
    \resizebox{0.9\linewidth}{!}{$
    \mathbf{u}_t=
    \begin{bmatrix}
        \begin{aligned}
    &0.0051 \Delta u - 0.95u_x(1.07u - 0.0065v - 0.17) + 0.98u_y(0.0045u - 1.01v + 0.17) + 0.053 \\
    &0.0051 \Delta v -0.82v_x(1.22u + 0.0078v - 0.18) - 0.91v_y(0.0063u + 1.08v - 0.17) + 0.058
        \end{aligned}
    \end{bmatrix} 
    $}
\end{equation}
which is very close to the ground truth of the governing PDE. Although the selection of candidate differential operators is crucial for identifying the genuine PDE, the $\Pi$-block shows better interpretability compared with the prolonged nested function formed by the FCNN or CNN. 

}

\end{document}